\documentclass[11pt]{article}

\usepackage[preprint]{acl}

\usepackage{times}
\usepackage{latexsym}
\usepackage{booktabs}
\usepackage{enumitem}
\usepackage{dblfloatfix}
\usepackage{placeins}
\usepackage{float}
\usepackage{svg}
\usepackage[T1]{fontenc}
\usepackage[utf8]{inputenc}
\usepackage{booktabs,tabularx,makecell,placeins}
\newcolumntype{Y}{>{\raggedright\arraybackslash}X}
\newcolumntype{C}{>{\centering\arraybackslash}X}
\usepackage{microtype}
\usepackage{caption}
\usepackage{xltabular}
\usepackage{caption}
\usepackage{inconsolata}
\usepackage{graphicx}
\usepackage{tikz}
\usetikzlibrary{arrows.meta, positioning}
\usepackage{amsmath}

\title{Learning Complementary Action Modeling\\
from Automotive Maintenance Instructions}

\author{
\textbf{Jiaqi Wu\textsuperscript{1,2,*}} \quad 
\textbf{Bai Li\textsuperscript{3,*}} \quad 
\textbf{Jochen Hartmann\textsuperscript{2}} \quad 
\textbf{Martin Gaedke\textsuperscript{3}} \quad 
\textbf{Sander Stuijk\textsuperscript{1}}
\\[0.3em]
\textsuperscript{1}Eindhoven University of Technology, Eindhoven, The Netherlands
\\
\textsuperscript{2}BMW Group, Munich, Germany
\\
\textsuperscript{3}Chemnitz University of Technology, Chemnitz, Germany
\\[0.3em]
\small{ \textsuperscript{*}Equal contribution. Correspondence: \texttt{jiaqi.wu@bmw.de}, \texttt{bai.li@informatik.tu-chemnitz.de} 
}
}

\begin{document}
\maketitle
\begin{abstract}
A minute lexical variation can reverse the procedural meaning of an instruction even when the rest of the sentence remains unchanged. In automotive maintenance instructions, this pattern often appears when an action phrase turns an instruction into its procedural counterpart. The entities, modifiers, and surrounding context remain largely invariant, while the action phrase determines the procedural relation. We define this task as Complementary Action Modeling (CAM). Given a maintenance instruction, the goal is to identify or generate its procedural counterpart by modifying the action phrase while preserving the remaining sentence context. This task focuses on three aspects: distinguishing complementarity from surface similarity, controlling generation at the action-phrase level, and evaluating relational correctness using retrieval, overlap-based, and human evaluation. Using a German automotive maintenance dataset, we examine these questions through candidate matching and controlled Seq2Seq generation. The results show that complementary maintenance instructions are best modeled as procedural associations grounded in subtle lexical cues. They should therefore not be treated as ordinary cases of sentence similarity or synonym-based paraphrasing.
\end{abstract}

\section{Introduction}

\begin{figure}[h]
    \centering
    \includegraphics[width=0.95\columnwidth]{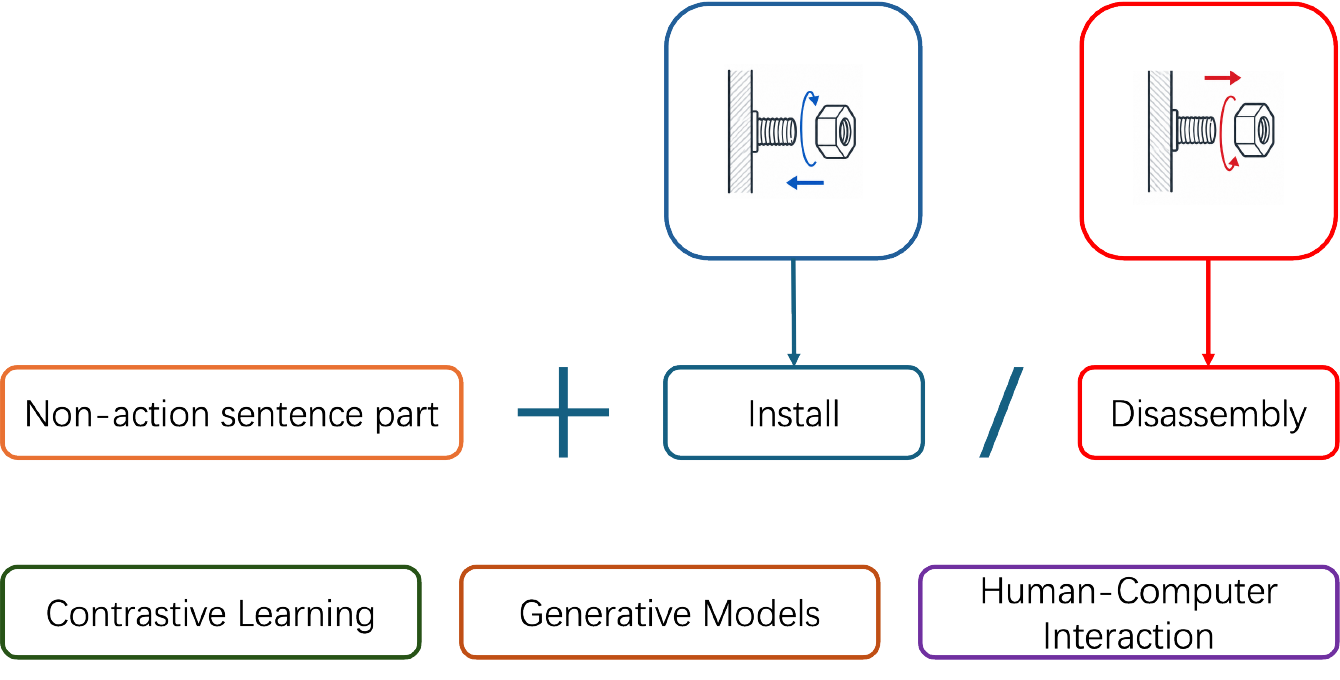}
    \caption{Illustration of Complementary Action Modeling (CAM). CAM preserves the non-action context while transforming the action phrase into its complementary counterpart.}
    \label{fig:cam}
\end{figure}

Minute lexical variations in maintenance instructions given to human operators and robots can determine the direction and objective of their actions. While the surrounding context may remain virtually unchanged, differences within the action phrases determine whether a specific component is to be installed or removed. Actual operational behaviors must be represented in a form that an intelligent system can reliably interpret. This capability is relevant to embodied intelligence and language-guided systems, which must distinguish similar actions and recognize complementary operations performed on the same object \citep{pmlr-v205-ichter23a,pmlr-v229-zitkovich23a}. We define Complementary Action Modeling (CAM) to address this problem. An illustration of CAM is shown in Fig.~\ref{fig:cam}.

Existing procedural text benchmarks do not fully capture this problem scenario. This is particularly evident in benchmark scenarios centered on procedural paragraphs and dynamic world tracking \citep{dalvi2018tracking}. However, automotive maintenance instructions are highly repetitive. Strict constraints on vocabulary usage often leave room for variation only within a very narrow segment. When the action phrase is substituted to alter the procedural relationship, the remainder of the sentence remains unchanged. This characteristic distinguishes CAM from a standard semantic similarity task; instructions that are most similar at the lexical level may, in terms of actual operation, be entirely incorrect. Nor is it a typical paraphrase generation task, as its objective is not to preserve the overall meaning of the sentence, but rather to achieve a controlled transformation of the operational relationship. Furthermore, it is not a contradiction detection task, as its focus lies on a complementary procedural correspondence embedded within the context of a single workflow, rather than on sentence-level incompatibility in a broader sense.

Another rationale for clearly defining CAM lies at the methodological level. Beyond evaluating neural models, this work follows a human-verified workflow for problem formalization, data construction, and result interpretation. Understanding the structure of maintenance procedures requires domain knowledge, which informs the rule-based alignment process used to formalize complementary relationships. Ambiguous actions then undergo selective human verification, while learning-based models are used to generalize beyond deterministic rule patterns. 

In this sense, CAM reflects a lightweight form of human-machine collaboration: Human expertise defines and verifies the procedural relation, while computational models scale this knowledge to larger sets of maintenance instructions. Trustworthy human judgment remains important for ensuring the reliability of downstream modeling pipelines \citep{bakaev2020distributionalgroundtruthnonredundant}.

CAM can also be viewed as an auxiliary modeling task for domain-specific procedural documents, where end users and domain experts may benefit from systems that make implicit procedural relations explicit \citep{esposito2023enduser}. In this setting, the contrastive model retrieves complementary action pairs that already exist within structured procedures, while the generative model tests whether complementary counterparts can be produced through controlled action-level generation.

The present study asks three research questions. (1) Can a model distinguish complementarity from surface similarity? This question is central because complementary instructions often share most of their tokens, making lexical overlap a poor proxy for the underlying procedural relation. (2) Can a generative model modify only the action phrase instead of freely rewriting the whole sentence? Fluent output is not sufficient: a generated sentence may look well-formed and still fail if it changes the correct context or preserves the source action. (3) How should CAM be evaluated? Overlap-based text generation metrics may reward outputs that remain close to the source even when they do not express the intended complementary relation. Relation-sensitive retrieval metrics and targeted human evaluation are therefore also needed. To answer these questions, we study CAM on a dataset derived from German automotive maintenance manuals,\footnote{The anonymized dataset and experimental resources will be publicly released for research purposes upon acceptance.} where automotive instructions are organized within hierarchical maintenance procedures. We examine the task through candidate matching and controlled sequence-to-sequence generation. Complementary maintenance instructions should be modeled as procedural associations grounded in subtle lexical cues rather than treated as ordinary instances of paraphrase or contradiction.

\section{Related Work}

CAM relates to three lines of work: procedural text understanding, industrial information extraction and semantic matching, and procedural generation. Unlike prior work, CAM focuses on complementary action relations in highly repetitive automotive maintenance instructions, where small lexical changes alter the procedural relation while most context remains unchanged.

Procedural text understanding models actions, entities, and state changes in instructional text. Existing benchmarks focus on tracking entity states in process descriptions~\cite{dalvi2018tracking}, with transformer-based methods further improving entity tracking~\cite{gupta2019effective}. Other work introduces richer structures, such as dynamic knowledge graphs and entity-action-location reasoning graphs~\cite{das2018building,huang2021reasoning}. These studies mainly target curated domains where state changes and procedure order are often explicit.

Industrial information extraction frequently uses rule-based or hybrid methods because they are interpretable and can encode expert knowledge. Rule-based systems remain useful in enterprise settings with limited supervision~\cite{chiticariu2013rule}, but are less robust to linguistic variation and implicit action semantics. Contrastive learning provides a data-driven alternative for semantic matching. InfoNCE-style objectives learn pairwise alignment from positive and negative examples~\cite{oord2018representation}, although random instance-level evaluation may overlook leakage in structurally repetitive documents~\cite{sogaard-etal-2021-need}.

Generative and script-based models treat procedural knowledge as action generation or event prediction. Neural checklist, state-aware, and pretrained language models have been used for coherent procedural generation~\cite{kiddon2016globally,dhingra-etal-2018-neural,H_Lee_2020}. Retrieval-augmented and script-based methods further incorporate external steps or event knowledge~\cite{nishimura2019procedural,sakaguchi2021script}. However, inverse or complementary action relations are usually addressed only indirectly. Broader event relation studies~\cite{zhou2022claretpretrainingcorrelationawarecontexttoevent,zhu2022generativeapproachscriptevent} suggest relevant directions but leave complementary action relations in automotive maintenance instructions underexplored.

\section{Dataset Construction}

This section describes the construction of a dataset for Complementary Action Modeling (CAM) from German automotive maintenance manuals. Our goal is to collect paired maintenance instructions while preserving the procedural structure. Automotive maintenance manuals describe goal-directed sequences of actions over objects, tools, and intermediate states, and their interpretation depends on contextual structure.

\subsection{Source Maintenance Manuals}

Our source documents are automotive maintenance manuals provided by an original equipment manufacturer (OEM). The manuals describe maintenance and repair procedures as structured textual instructions organized hierarchically into repair tasks, repair processes, and repair steps.

A repair task specifies a high-level maintenance objective. Each task is decomposed into multiple repair processes, and each process consists of fine-grained repair steps expressed as short imperative instructions. In this work, we operate at the level of individual repair steps, but we retain process membership as an essential contextual variable. Complementary relations in maintenance manuals arise within coherent procedural contexts such as disassembly and reassembly workflows.

To preserve this structure, we represent each repair process as a \emph{bucket} and assign all steps from the same process the same \emph{bucket identifier}. The bucket structure preserves the local procedural context and later supports process-aware evaluation splits.

\subsection{Rule-Based Construction of Complementary Action Pairs}

We construct complementary action pairs with a two-stage rule-based alignment pipeline followed by manual verification. The first stage operates at the process level and identifies candidate process pairs that are likely to stand in a complementary relation, such as disassembly versus assembly process. The second stage operates within each matched process pair and aligns individual repair steps.

At the core of the rule matching process is a curated German lexicon of complementary action expressions. Candidate pairs are first identified based on compatible action opposition, such as install/remove, connect/disconnect, or tighten/loosen. After detecting candidate action pairs, the matched action expressions are removed and the remaining object- and context-bearing text is referred to as the action-stripped skeleton, which is then compared across candidate pairs. Candidate pairs are retained when their action-stripped skeletons match exactly or differ only by conservative surface variations.

To handle such variation, the rule matching process combines approximate string matching with whitelist-style normalization rules. These rules tolerate harmless differences in word order, light modifiers, and synonym substitutions, such as \emph{vorsichtig} ("carefully") or \emph{wie abgebildet} ("as illustrated"). At the same time, the rule matching process applies hard procedural constraints to avoid spurious alignments. In particular, side- and location-sensitive expressions are treated as incompatibility signals, so that otherwise similar instructions are not matched if they refer to different physical positions.

At the step level, the pipeline additionally exploits shared illustration references, which are particularly useful in highly repetitive repair subprocedures where multiple steps differ only in localized object references.

Although the rule-based pipeline can detect certain one-to-many correspondences, such as grouped screw-removal operations that correspond to several later fastening steps, these cases are excluded from the final learning benchmark. We retain only high-confidence one-to-one complementary pairs in order to maintain an unambiguous retrieval and generation setting for downstream learning-based experiments.

\subsection{Manual Verification and Provenance}
Manual verification is applied selectively rather than to all automatically proposed pairs. High-confidence cases include strict rule matches and whitelist-supported matches. Candidates that cannot be resolved by these deterministic rules but remain plausible under looser approximate matching, are exported for human review, where ambiguous candidates are confirmed or rejected.

We retain provenance labels indicating whether a pair was obtained through strict rule matching, whitelist-supported normalization, or approximate matching with manual confirmation. This provenance distinguishes rule-resolved cases from more ambiguous semantic cases. In our corpus-level audit, the rule-based matching pipeline resolves approximately 59\% of the final complementary pairs before manual intervention, while the remaining retained pairs require human adjudication due to lexical divergence, implicit component references, or context-dependent formulations.

This distinction enables a natural separation between rule-resolved and rule-unresolved cases. Rule-resolved cases are primarily governed by explicit lexical and structural regularities, whereas rule-unresolved cases represent semantically harder instances that cannot be recovered reliably through deterministic heuristics alone.

After deduplication and filtering for high-confidence one-to-one alignments, the final benchmark contains 1,459 complementary instruction pairs. Each pair is associated with its original repair-process bucket and provenance label. These annotations later support process-aware evaluation splits and rule-resolved versus rule-unresolved analysis.

The rule-based pipeline serves two purposes in this work: it constructs a high-confidence CAM dataset while also separating rule-resolved cases from semantically harder ones. The learning-based models are motivated by the latter, where surface similarity, local lexical opposition, and handcrafted constraints are not sufficient to recover the correct counterpart.

\section{Experimental Design}
We evaluate CAM under two experimental paradigms: candidate matching and controlled generation. Candidate matching formulates CAM as a relation-aware retrieval problem, whereas controlled generation formulates it as a conditional sequence-to-sequence generation problem. The former evaluates whether a model can retrieve the correct counterpart from a candidate set, while the latter evaluates whether it can generate a complementary instruction under action-level control.

\subsection{Candidate Matching Experiment}
\label{sec:candidate matching experiment}
In the candidate matching experiment, the model receives a source maintenance instruction and selects the corresponding procedural counterpart from a set of candidates. This experiment primarily addresses the first research question: \emph{Can a model distinguish complementarity from surface similarity?} 

In this experiment, each sample consists of a source instruction together with a candidate pool containing one correct procedural counterpart and other non-matching instructions from the same evaluation split.

Given a source instruction $x_i$ and a candidate set $C=\{c_1, c_2, \ldots, c_M\}$, the objective is to identify the procedural counterpart $c_i^+$ of $x_i$. The model learns an encoding function $f_\theta(\cdot)$ that maps the source instruction and each candidate instruction to vector representations:
\begin{equation}
z_x = f_\theta(x), \quad z_c = f_\theta(c)
\end{equation}

Matching scores are then computed using vector similarity:
\begin{equation}
s(x,c) = z_x^\top z_c
\end{equation}

Candidates are ranked by this score, with the gold counterpart expected to receive the highest rank.

\subsubsection{Training Objective}
We train the model with a symmetric InfoNCE loss using verified CAM pairs as positives. Within each batch, all non-matching target instructions naturally serve as in-batch negatives. For example, \emph{remove the protective cover} and \emph{install the protective cover} form a positive CAM pair. For a batch containing $N$ positive sample pairs:
\begin{equation}
\{(x_i, y_i)\}_{i=1}^{N}
\end{equation}

where $x_i$ is the source instruction and $y_i$ is its procedural counterpart. The model encodes the source and target separately:
\begin{equation}
z_i^x = f_\theta(x_i), \qquad z_i^y = f_\theta(y_i)
\end{equation}

It then computes a similarity matrix across all source-target pairs:
\begin{equation}
S_{ij} = \frac{(z_i^x)^\top z_j^y}{\tau}
\end{equation}

Here, $\tau$ denotes the temperature parameter.

For each $x_i$, the correct match is $y_i$, while all other targets $y_j$ within the batch serve as negative samples. The corresponding loss function can be formulated as:
\begin{equation}
\mathcal{L}_{x \to y}
= -\frac{1}{N}
\sum_{i=1}^{N}
\log
\frac{\exp(S_{ii})}
{\sum_{j=1}^{N} \exp(S_{ij})}
\end{equation}

We apply the same objective in the reverse direction:
\begin{equation}
\mathcal{L}_{y \to x}
= -\frac{1}{N}
\sum_{i=1}^{N}
\log
\frac{\exp(S_{ii})}
{\sum_{j=1}^{N} \exp(S_{ji})}
\end{equation}

The final symmetric contrastive loss is then given by:
\begin{equation}
\mathcal{L}_{\mathrm{match}}
=
\frac{1}{2}
\left(
\mathcal{L}_{x \to y}
+
\mathcal{L}_{y \to x}
\right)
\end{equation}

\subsubsection{Experiment Setup}
We use a bi-encoder architecture for the candidate matching experiment. Source and candidate instructions are encoded independently and ranked by dot-product similarity. This architecture supports efficient retrieval over large candidate sets. We compare four encoders, summarized in Table \ref{tab:encoder-families}: mBERT, German BERT, multilingual Sentence-Transformer, and a character-level Transformer. Detailed encoder configurations are provided in Appendix ~\ref{app:reproducibility_details}.

\begin{table}[h]
\centering
\small
\setlength{\tabcolsep}{3pt}
\begin{tabularx}{\columnwidth}{p{2.3cm}p{4.9cm}}
\toprule
\textbf{Encoder} & \textbf{Description} \\
\midrule
mBERT & Multilingual Transformer baseline \\
German BERT & German-specific pretrained encoder \\
Multilingual ST & Sentence-transformer retrieval encoder \\
Char Transformer & Character-level encoder trained from scratch \\
\bottomrule
\end{tabularx}
\caption{Encoders used in candidate matching.}
\label{tab:encoder-families}
\end{table}

\subsection{Controlled Generation Experiment}
In the controlled generation experiment, the model receives a source instruction and generates its complementary counterpart. The output should modify only the action phrase while preserving the surrounding context. This experiment primarily addresses the second research question: \emph{Can a generative model modify only the action phrase instead of freely rewriting the whole sentence?}

Given a source maintenance instruction $x$, the model generates a target instruction $y$ that expresses its procedural counterpart: 
\begin{equation}
P_\theta(y \mid x)
\end{equation}

Thus, CAM generation requires both action transformation and context preservation. 

\subsubsection{Training Objective}
The primary training objective combines bidirectional sequence-to-sequence loss with an embedding-space contrastive regularization term. For a training pair $(x,y)$, the model maximizes the probability of the target sequence conditioned on the input:
\begin{equation}
P_\theta(y|x) = \prod_{t=1}^T P_\theta(y_t | y_{<t}, x)
\end{equation}

The corresponding negative log-likelihood loss is:
\begin{equation}
L_{NLL} = -\sum_{t=1}^T \log P_\theta(y_t | y_{<t}, x)
\end{equation}

Since token-level loss alone may emphasize surface reproduction, we additionally use an embedding-space contrastive regularizer in the main generation setting. The contrastive regularizer encourages representations of paired source and target instructions to remain close while separating non-matching targets in representation space. This helps the generative model learn not only token-level output, but also sentence-level procedural associations.

The final loss is:
\begin{equation}
\mathcal{L} = \mathcal{L}_{gen} + \lambda \mathcal{L}_{ctr},
\end{equation}
where $\mathcal{L}_{ctr}$ is the embedding-space contrastive loss and $\lambda$ controls its weight.

\subsubsection{Experimental Setup}
In our experiment, we use pretrained encoder-decoder models to map a source instruction to its complementary target. We use a row-wise 80/10/10 train/validation/test split. Each pair is used in both directions, so the model is trained to generate $y$ from $x$ and $x$ from $y$. 

We evaluate three multilingual Seq2Seq backbone families, summarized in Table~\ref{tab:seq2seq-backbones}, to compare the effect of pretraining and model capacity. Decoding uses greedy generation with a maximum length of 128 tokens. The best checkpoint is selected by validation ROUGE-L. Additional reproducibility details and evaluation implementations are summarized in Appendix ~\ref{app:reproducibility_details}.

\begin{table}[h]
\centering
\small
\setlength{\tabcolsep}{3pt}
\begin{tabularx}{\columnwidth}{p{1.5cm}p{5.8cm}}
\toprule
\textbf{Backbone} & \textbf{Description} \\
\midrule
mBART & Multilingual encoder-decoder model for sequence-to-sequence generation. \\
mT5 & Multilingual T5 model used to test capacity effects across model sizes. \\
Flan-T5 & Instruction-tuned T5 variant used as a generation baseline. \\
\bottomrule
\end{tabularx}
\caption{Seq2Seq backbones used in controlled generation.}
\label{tab:seq2seq-backbones}
\end{table}

\begin{table*}[t]
\centering
\small
\begin{tabularx}{\textwidth}{lCCCC}
\toprule
\textbf{Encoder} & \textbf{R@1} & \textbf{R@5} & \textbf{R@10} & \textbf{MRR} \\
\midrule
Char Transformer & $0.7134 \pm 0.0313$ & $0.9445 \pm 0.0271$ & $0.9857 \pm 0.0133$ & $0.8083 \pm 0.0227$ \\
mBERT & $0.7415 \pm 0.0168$ & $\mathbf{0.9612 \pm 0.0259}$ & $\mathbf{0.9936 \pm 0.0072}$ & $\mathbf{0.8306 \pm 0.0152}$ \\
German BERT & $0.7321 \pm 0.0275$ & $0.9588 \pm 0.0205$ & $0.9924 \pm 0.0098$ & $0.8234 \pm 0.0197$ \\
Multilingual ST & $\mathbf{0.7417 \pm 0.0216}$ & $0.9579 \pm 0.0211$ & $0.9935 \pm 0.0053$ & $0.8304 \pm 0.0163$ \\
\bottomrule
\end{tabularx}
\caption{Candidate matching results under the \texttt{groups\_native} setting. Scores are reported as mean $\pm$ standard deviation over 10 folds.}
\label{tab:candidate-matching}
\end{table*}

\section{Evaluation and Results}
We first describe the evaluation protocol and then report results for candidate matching, controlled generation, and additional diagnostic analyses.

\subsection{Evaluation Protocol}
We evaluate CAM under two experimental paradigms: candidate matching and controlled generation. The evaluation protocol is designed to measure both retrieval accuracy and action-level transformation quality.

\subsubsection{Candidate Matching Evaluation}
\label{subsubsec:5.1.1_candidate_evaluation}
For candidate matching, we consider three split settings: \texttt{pairs}, \texttt{groups}, and \texttt{groups\_native}. 

The \texttt{pairs} setting performs row-wise splitting over all complementary instruction pairs, regardless of their originating repair process. The \texttt{groups} setting uses bucket-level splitting on the rebucketed dataset, while \texttt{groups\_native} restricts this evaluation to the original repair-process buckets derived from the maintenance manual hierarchy. This prevents highly similar process-local templates from appearing across training and evaluation splits.

Candidate matching is evaluated with 10-fold cross-validation. In each fold, eight folds are used for training, one for validation, and one for testing. 

We report Recall@1, Recall@5, Recall@10, and MRR. Recall@K measures whether the correct counterpart appears within the top-K retrieved candidates, while MRR reflects the average ranking quality across the full candidate list.

We focus the main text on the conservative \texttt{groups\_native} setting and report the full comparison in Appendix~\ref{app:contrastive_retrieval_results}.

\subsubsection{Controlled Generation Evaluation}

For controlled generation, the best checkpoint is selected by validation ROUGE-L and final automatic metrics are reported on the held-out test split.

We report BLEU and ROUGE as automatic overlap-based metrics. BLEU measures n-gram overlap between generated and reference counterparts, while ROUGE evaluates lexical and contextual similarity.

However, overlap-based metrics alone do not determine whether the action phrase has been transformed correctly. We therefore additionally conduct human evaluation on 100 directional outputs from the held-out test split, stratified into 50 rule-resolved and 50 rule-unresolved cases.

The human evaluation focuses on semantic complementarity and component consistency. Semantic complementarity measures whether the generated instruction expresses the correct complementary action, while component consistency evaluates whether the generated output preserves the correct procedural entity or component.

\subsubsection{Rule-Resolved and Rule-Unresolved Analysis}

We additionally report candidate matching results separately for rule-resolved and rule-unresolved pairs. Rule-resolved pairs are obtained by strict or whitelist-supported matching, while rule-unresolved pairs correspond to manually confirmed matches not directly resolved by deterministic rules.

This split allows us to examine whether learned models remain effective on more challenging cases beyond deterministic rule patterns.

\subsection{Candidate Matching Results}
Table~\ref{tab:candidate-matching} reports candidate matching results on the conservative \texttt{groups\_native} setting.

The results show that CAM is learnable as a closed-set matching task: the strongest systems reach approximately 0.74 R@1 and 0.83 MRR. The multilingual sentence-transformer achieves the best R@1, while mBERT achieves the strongest Recall@5/10 and the best overall MRR. This difference suggests that CAM performance should not be reduced to a single retrieval score.

Transformer-based encoders outperform the character-level encoder, although the latter remains competitive. This suggests that surface lexical information is useful for CAM, but pretrained multilingual and German-specific representations provide stronger signals for identifying procedural counterparts, even in highly formulaic maintenance instructions. Extended retrieval results across all split settings are reported in Appendix~\ref{app:contrastive_retrieval_results}.

We further analyze the held-out evaluation results separately on rule-resolved and rule-unresolved subsets, as shown in Table~\ref{tab:rule_split_results}. As expected, rule-resolved pairs are easier, but performance on rule-unresolved cases remains strong, with the multilingual Sentence-Transformer achieving over 0.82 Recall@1. This indicates that learned representations capture complementary relations beyond deterministic rule patterns.

\begin{table}[H]
\centering
\small
\setlength{\tabcolsep}{5pt}
\begin{tabularx}{\columnwidth}{lCCCC}
\toprule
Subset & R@1 & R@5 & R@10 & MRR \\
\midrule
Rule-resolved & 0.9542 & 0.9977 & 0.9989 & 0.9739 \\
Rule-unresolved & 0.8237 & 0.9710 & 0.9851 & 0.8856 \\
\bottomrule
\end{tabularx}
\caption{Candidate matching results on rule-resolved and rule-unresolved subsets using the multilingual Sentence-Transformer under the \texttt{groups\_native} protocol.}
\label{tab:rule_split_results}
\end{table}

\subsection{Controlled Generation Results}
\label{subsec:generation-results}

Table~\ref{tab:generation-results} reports controlled generation results for multilingual Seq2Seq backbones trained under the bidirectional full-sentence generation setup.

\begin{table}[H]
\centering
\small
\setlength{\tabcolsep}{4pt}
\begin{tabularx}{\columnwidth}{lCCCC}
\toprule
\textbf{Model} & \textbf{BLEU} & \textbf{R-1} & \textbf{R-2} & \textbf{R-L} \\
\midrule
mBART-large-50 & \textbf{62.9603} & \textbf{0.8083} & \textbf{0.7177} & \textbf{0.7970} \\
Flan-T5-large & 61.7858 & 0.8075 & 0.7133 & 0.7955 \\
mT5-large & 61.4361 & 0.7953 & 0.6968 & 0.7842 \\
Flan-T5-base & 57.8672 & 0.7860 & 0.6758 & 0.7729 \\
mT5-base & 48.0914 & 0.7128 & 0.5837 & 0.6979 \\
mT5-small & 0.1488 & 0.0174 & 0.0000 & 0.0169 \\
\bottomrule
\end{tabularx}
\caption{Controlled generation results on the held-out test set. The best checkpoint is selected by validation ROUGE-L. All models are trained with bidirectional generation and embedding-space contrastive regularization ($\lambda=0.5$, $\tau=0.05$). R-1, R-2, and R-L denote ROUGE-1, ROUGE-2, and ROUGE-L, respectively.}
\label{tab:generation-results}
\end{table}

The generation results show that strong pretrained multilingual Seq2Seq models can produce outputs with high reference overlap. The best checkpoint is selected by validation ROUGE-L, and final results are reported on the held-out test set. mBART-large-50 obtains the strongest test scores, reaching 62.9603 BLEU and 0.7970 ROUGE-L. This suggests that much of the surrounding procedural context can be preserved in the generated counterpart.

However, overlap-based metrics alone do not prove CAM success. In CAM generation, copying is both necessary and risky: the model should preserve the component and context, but must still transform the action phrase. We therefore complement BLEU and ROUGE with human evaluation of semantic complementarity and component consistency.

Among 100 directional test outputs, 95 were judged semantically complementary and 5 were incorrect with respect to the intended action transformation. All sampled outputs preserved the relevant component or procedural entity, suggesting that remaining errors mainly concern action-level transformation rather than component drift.

In Table~\ref{tab:human-eval-generation}, Correct and Incorrect refer to semantic complementarity judgments. Component consistency is not shown as a separate column because all sampled outputs preserved the relevant component or procedural entity.

\begin{table}[H]
\centering
\small
\setlength{\tabcolsep}{10pt}
\begin{tabularx}{\columnwidth}{lCC}
\toprule
Subset & Correct & Incorrect \\
\midrule
Rule-resolved & 49 & 1 \\
Rule-unresolved & 46 & 4 \\
\bottomrule
\end{tabularx}
\caption{Human evaluation of semantic complementarity for generated outputs by rule-resolved and rule-unresolved subset.}
\label{tab:human-eval-generation}
\end{table}

These findings show that high overlap-based generation scores should still be interpreted together with action-level semantic judgments. Model capacity and pretraining quality also strongly affect generation performance. mBART-large-50 performs best, while large T5 variants remain competitive. In contrast, mT5-small degenerates under the same training setting, suggesting that CAM generation is more sensitive to model capacity and optimization stability than candidate matching.

\subsection{Cross-Vehicle Generalization Case Study}
\label{subsec:cross-vehicle}

Beyond the controlled benchmark, we further examine whether CAM generalizes to maintenance instruction data from an unseen vehicle platform by applying the best-performing candidate matching checkpoint in inference mode only, without additional fine-tuning. The new vehicle data is not used during training, validation, or testing in the main benchmark.

Because the new vehicle data does not contain a fully annotated benchmark, we evaluate the retrieved counterparts through manual verification. A prediction is counted as accepted if it expresses a semantically valid complementary action relation for the source instruction.

Among 1,584 reviewed inference results, 1,022 predictions are accepted as valid complementary matches, corresponding to an acceptance rate of 64.5\%.

The cross-vehicle case study suggests that CAM captures procedural complementary relations that transfer beyond the original vehicle-specific benchmark. Although the evaluation is based on manual verification rather than a fully annotated retrieval benchmark, the results indicate promising generalization behavior on unseen automotive maintenance documentation.

\subsection{Discussion}
\label{subsec:mixed-evaluation}

Taken together, the benchmark results, human evaluation, and cross-vehicle case study show that CAM requires evaluation from multiple perspectives. In this work, mixed evaluation refers to combining retrieval metrics for counterpart identification, overlap-based metrics for contextual adequacy, and human evaluation for action-level correctness. The combined results show that candidate matching and controlled generation expose different aspects of CAM. Candidate matching evaluates whether the correct counterpart can be identified in a closed candidate set, while generation tests whether the model can produce a complementary instruction. Human evaluation shows that generated outputs usually preserve the component context reliably, while the remaining errors mainly concern action-level transformation. Together with the rule-resolved and rule-unresolved analysis, these results suggest that CAM is learnable but remains challenging when complementarity cannot be resolved by deterministic lexical patterns alone.

\section{Conclusion}
This paper introduced Complementary Action Modeling (CAM) for automotive maintenance instructions, where the goal is to identify or generate a procedural counterpart by transforming the action phrase while preserving the surrounding context. Our results show that CAM is learnable as a closed-set matching task, with strong retrieval performance across multilingual and German-specific encoders in the \texttt{groups\_native} setting. The rule-resolved and rule-unresolved analysis further shows that cases beyond deterministic matching rules remain harder but still learnable.

For controlled generation, mBART-large-50 is the strongest backbone, achieving the best BLEU and ROUGE-L scores among the evaluated models. Human evaluation shows that generated outputs preserve component context reliably, while the remaining errors mainly concern action-level transformation. Together, these findings support mixed evaluation for CAM: retrieval metrics assess counterpart identification, overlap-based metrics capture contextual adequacy, and human evaluation verifies action-level correctness.

Overall, complementary maintenance instructions should be modeled as procedural associations grounded in subtle lexical cues rather than as ordinary sentence similarity or paraphrase. Future work should extend CAM to broader maintenance domains, develop automatic action-aware diagnostics, and study more challenging cross-document and cross-vehicle generalization settings, including error analysis of no-op, wrong-action, context-drift, and near-match failures. In short, CAM succeeds only when a model identifies or generates the right counterpart, preserves the context, and changes the action correctly.

\section*{Limitations}

While our results demonstrate the effectiveness of learning-based approaches for modeling complementary actions in automotive maintenance instructions, several limitations remain.

First, the dataset is derived primarily from German automotive maintenance instructions from a single OEM, which may limit generalization to other manufacturers, languages, or manual styles. Although the proposed methods are not tied to a specific action lexicon, broader validation across domains remains necessary.

Second, the current benchmark depends on an initial rule-based alignment  stage followed by manual verification. Although learning substantially improves robustness beyond handcrafted rules, biases or constraints introduced during pair construction may still affect the learned representations. Fully data-driven discovery of complementary actions therefore remains an open challenge.

Third, the formulation emphasizes locally constrained disambiguation within structurally related procedural contexts rather than global retrieval across entire maintenance instructions. While this reflects common industrial authoring practices, it does not address cases where complementary actions occur in distant or weakly connected sections.

Fourth, the generative formulation exhibits a strong dependence on model capacity and pretraining quality. Large pretrained sequence-to-sequence models perform reliably, whereas low-capacity models may show instability or degenerate generation, suggesting that purely generative approaches may be unsuitable for constrained model settings and motivating future work on parameter-efficient or hybrid contrastive–generative methods.

Finally, the cross-vehicle generalization analysis is based on manual verification of inference outputs rather than a fully annotated retrieval benchmark. It provides evidence of transfer to unseen maintenance documentation, but does not replace a controlled benchmark with exhaustive gold counterparts.

\section*{Acknowledgments}

The authors acknowledge the use of AI-assisted tools for language refinement and editorial support during manuscript preparation. All technical content, experimental design, analysis, and conclusions were verified and approved by the authors.

\bibliography{custom}

\appendix

\section{Reproducibility Details}
\label{app:reproducibility_details}
This appendix summarizes additional implementation and reproducibility details for the CAM experiments, including evaluation protocols, metric implementations, computational resources, training configurations, and extended retrieval results.

\subsection{Evaluation Metrics and Implementations}
Retrieval evaluation uses Recall@1, Recall@5, Recall@10, and Mean Reciprocal Rank (MRR) computed from ranked candidate lists produced by the bi-encoder retrieval models.

Generation evaluation uses BLEU and ROUGE-1/2/L. BLEU is computed with SacreBLEU, and ROUGE scores are computed with the HuggingFace \texttt{evaluate} implementation using standard settings.

For retrieval experiments, reported values correspond to mean$\pm$std across 10 folds. For controlled generation, the best checkpoint is selected by validation ROUGE-L and final results are reported on the held-out test split.

\subsection{Computational Resources}

All experiments were conducted on a single NVIDIA A100 GPU. The overall training budget across retrieval and generation experiments remained below approximately 100 GPU hours.

We use publicly available pretrained backbones and fine-tune them under the experimental configurations reported in the following appendix sections.

\section{Contrastive Retrieval Configuration}
\label{app:contrastive_config}
This section summarizes the training configurations and encoder settings used in the contrastive retrieval experiments described in Section~\ref{sec:candidate matching experiment}. The appendix reports both the shared default configuration and the experiment-specific overrides used for the \texttt{pairs}, \texttt{groups}, and \texttt{groups\_native} evaluation settings. Table~\ref{tab:contrastive_common} lists the default hyperparameters shared across experiments, Table~\ref{tab:contrastive_overrides} summarizes the setting-specific overrides, and Table~\ref{tab:contrastive_encoders} reports the encoder architectures, pooling strategies, and normalization choices used in retrieval.

\subsection{Common Settings}
\label{sec:app:Common Settings}

\begin{table}[H]
\centering
\small
\setlength{\tabcolsep}{6pt}
\begin{tabularx}{\columnwidth}{lY}
\toprule
Setting & Value \\
\midrule
Random seed & 42 \\
Device & auto \\
Optimizer & AdamW \\
Learning rate & $1\times 10^{-5}$ \\
Weight decay & 0.01 \\
Epochs (default) & 20 \\
Batch size (default) & 64 \\
Max length & 256 \\
Grad clip norm & 1.0 \\
Log interval & 10 steps \\
Mixed precision & True (CUDA only) \\
InfoNCE temperature & 0.05 \\
Eval metrics & R@1, R@5, R@10, MRR \\
Text columns & \texttt{aus\_text}, \texttt{ein\_text} \\
Bucket column & \texttt{bucket\_id} \\
DataLoader workers & 0 \\
\bottomrule
\end{tabularx}
\caption{Common hyperparameters used across contrastive experiments
(from \texttt{configs/base.yaml}).}
\label{tab:contrastive_common}
\end{table}

\clearpage
\subsection{Experiment-Specific Overrides}
\label{sec:app:exp_overrides}
\noindent\makebox[\textwidth]{%
\begin{minipage}{\textwidth}
\centering
\small
\setlength{\tabcolsep}{4pt}

\begin{tabularx}{\textwidth}{lYYY}
\toprule
Setting & Pairs & Groups & Groups-native \\
\midrule
Dataset &
\makecell[l]{\texttt{train\_dedup\_by\_text}\\\texttt{\_pairs.csv}} &
\makecell[l]{\texttt{rebucketed\_singletons.csv}} &
\makecell[l]{\texttt{rebucketed\_singletons.csv}} \\
Split type & k-fold (row) & k-fold (group) & k-fold (group) \\
Group column & -- & \texttt{bucket\_id} & \texttt{bucket\_id} \\
$k$ (folds) & 10 & 10 & 10 \\
Fold index & 0 & 0 & 0 \\
Test fold offset & 1 & 1 & 1 \\
Epochs & 20 & 30 & 30 \\
Batch size & 64 & 128 & 128 \\
Temperature $\tau$ & 0.07& 0.05 & 0.05 \\
Native-only filter & -- & -- &
\makecell[l]{\texttt{bucket\_id} not starting\\with \texttt{``new''}} \\
\bottomrule
\end{tabularx}

\vspace{-0.5em}
\captionof{table}{Experiment-specific configuration overrides
(from \texttt{configs/pairs.yaml}, \texttt{configs/groups.yaml},
and \texttt{configs/groups\_native.yaml}).}
\label{tab:contrastive_overrides}
\end{minipage}%
}

\subsection{Encoder Configurations}
\label{sec:app:encoder_configurations}

\noindent\makebox[\textwidth]{%
\begin{minipage}{\textwidth}
\centering
\small
\setlength{\tabcolsep}{5pt}

\begin{tabularx}{\textwidth}{l l Y l}
\toprule
Encoder & Backend & Model / Architecture & Pooling / Normalize \\
\midrule
Multilingual BERT & HF & \texttt{bert-base-multilingual-cased} & CLS / L2-norm \\
German BERT & HF & \texttt{bert-base-german-cased} & CLS / L2-norm \\
ST-MPNet & ST & \texttt{paraphrase-multilingual-mpnet-base-v2} & ST / L2-norm \\
Char Transformer & Char &
\makecell[l]{V=5000, $d$=512, $h$=8, $L$=6,\\ FF=2048, drop=0.1} &
Mean / L2-norm \\
\bottomrule
\end{tabularx}

\vspace{-0.5em}
\captionof{table}{Configurations of different encoders used in contrastive experiments
(from \texttt{configs/encoders/*.yaml}).}
\label{tab:contrastive_encoders}
\end{minipage}%
}

\section{Contrastive Retrieval Results}
\label{app:contrastive_retrieval_results}
This section reports the full retrieval results across all split settings discussed in Section~\ref{subsubsec:5.1.1_candidate_evaluation}, including the \texttt{pairs}, \texttt{groups}, and \texttt{groups\_native} protocols with progressively stricter separation constraints. All reported values correspond to mean$\pm$std across 10-fold cross-validation.
\\

\noindent\makebox[\textwidth]{%
\begin{minipage}{\textwidth}
\centering
\small
\setlength{\tabcolsep}{6pt}

\begin{tabularx}{\textwidth}{lCCCC}
\toprule
\multicolumn{5}{l}{\textbf{Pairs} (k=10, mean$\pm$std)}\\
Encoder & Recall@1 & Recall@5 & Recall@10 & MRR \\
\midrule
Multilingual BERT & $0.9027\pm0.0250$ & $0.9863\pm0.0115$ & $0.9945\pm0.0060$ & $0.9388\pm0.0166$ \\
German BERT       & $0.8979\pm0.0206$ & $0.9808\pm0.0114$ & $0.9925\pm0.0048$ & $0.9348\pm0.0157$ \\
ST-MPNet          & $0.9089\pm0.0210$ & $0.9870\pm0.0112$ & $0.9966\pm0.0063$ & $0.9430\pm0.0135$ \\
Char Transformer  & $0.8472\pm0.0328$ & $0.9548\pm0.0157$ & $0.9726\pm0.0150$ & $0.8942\pm0.0233$ \\

\midrule
\multicolumn{5}{l}{\textbf{Groups} (k=10, mean$\pm$std)}\\
Encoder & Recall@1 & Recall@5 & Recall@10 & MRR \\
\midrule
Multilingual BERT & $0.6820\pm0.0204$ & $0.8680\pm0.0181$ & $0.9326\pm0.0172$ & $0.7667\pm0.0174$ \\
German BERT       & $0.6812\pm0.0214$ & $0.8672\pm0.0199$ & $0.9296\pm0.0184$ & $0.7672\pm0.0186$ \\
ST-MPNet          & $0.6785\pm0.0207$ & $0.8698\pm0.0203$ & $0.9345\pm0.0181$ & $0.7662\pm0.0174$ \\
Char Transformer  & $0.6656\pm0.0267$ & $0.8546\pm0.0197$ & $0.9203\pm0.0197$ & $0.7543\pm0.0218$ \\

\midrule
\multicolumn{5}{l}{\textbf{Groups-native} (k=10, mean$\pm$std)}\\
Encoder & Recall@1 & Recall@5 & Recall@10 & MRR \\
\midrule
Multilingual BERT & $0.7337\pm0.0198$ & $0.9599\pm0.0265$ & $0.9935\pm0.0072$ & $0.8259\pm0.0161$ \\
German BERT       & $0.7330\pm0.0243$ & $0.9619\pm0.0220$ & $0.9946\pm0.0073$ & $0.8260\pm0.0189$ \\
ST-MPNet          & $0.7384\pm0.0221$ & $0.9589\pm0.0199$ & $0.9935\pm0.0053$ & $0.8289\pm0.0166$ \\
Char Transformer  & $0.7058\pm0.0323$ & $0.9403\pm0.0248$ & $0.9867\pm0.0158$ & $0.8040\pm0.0244$ \\
\bottomrule
\end{tabularx}

\vspace{-0.5em}
\captionof{table}{Contrastive retrieval results for \texttt{pairs}, \texttt{groups}, and \texttt{groups-native}. All scores are mean$\pm$std over 10-fold cross-validation. Within each fold, the best checkpoint is selected by validation MRR; Recall@1/5/10 and MRR are reported on the validation split.}
\label{tab:contrastive_results}
\end{minipage}%
}

\end{document}